\documentclass[sigconf]{acmart}

\usepackage{booktabs}
\usepackage{amsmath}
\usepackage{graphicx}
\usepackage{xcolor}
\usepackage{algorithm}
\usepackage{algpseudocode}
\usepackage{float}
\usepackage{tikz}
\usetikzlibrary{arrows.meta,positioning,shapes.geometric}

\renewcommand\footnotetextcopyrightpermission[1]{}
\AtBeginDocument{%
  \setlength{\abovedisplayskip}{4pt plus 1pt minus 1pt}%
  \setlength{\belowdisplayskip}{4pt plus 1pt minus 1pt}%
  \setlength{\abovedisplayshortskip}{3pt plus 1pt minus 1pt}%
  \setlength{\belowdisplayshortskip}{3pt plus 1pt minus 1pt}%
}

\acmConference[Agent4IR @ KDD 2026]{KDD 2026 Workshop on AI Agents for Information Retrieval}{August 2026}{}
\acmYear{2026}
\copyrightyear{2026}
\setcopyright{none}

\begin{document}

\title[Contrastive Reflection]{Contrastive Reflection for Iterative Prompt Optimization}

\author{Derek Koh}
\affiliation{\institution{LinkedIn}\city{Sunnyvale}\state{California}\country{USA}}
\email{dkoh@linkedin.com}

\author{Jinghui Mo}
\affiliation{\institution{LinkedIn}\city{Sunnyvale}\state{California}\country{USA}}
\email{jmo@linkedin.com}

\author{Benjamin H. Le}
\affiliation{\institution{LinkedIn}\city{Sunnyvale}\state{California}\country{USA}}
\email{ble@linkedin.com}

\author{Jiening Zhan}
\affiliation{\institution{LinkedIn}\city{Sunnyvale}\state{California}\country{USA}}
\email{jzhan@linkedin.com}

\author{Baofen Zheng}
\affiliation{\institution{LinkedIn}\city{Sunnyvale}\state{California}\country{USA}}
\email{bzheng@linkedin.com}

\author{Kevin Bevis}
\affiliation{\institution{LinkedIn}\city{Sunnyvale}\state{California}\country{USA}}
\email{kbevis@linkedin.com}

\author{Nathaniel C. Owen}
\affiliation{\institution{LinkedIn}\city{Sunnyvale}\state{California}\country{USA}}
\email{chowen@linkedin.com}

\author{Lauren Elizabeth Charney}
\affiliation{\institution{LinkedIn}\city{Sunnyvale}\state{California}\country{USA}}
\email{lcharney@linkedin.com}

\author{Wenqiong Liu}
\affiliation{\institution{LinkedIn}\city{Sunnyvale}\state{California}\country{USA}}
\email{ecliu@linkedin.com}

\author{Jingwei Wu}
\affiliation{\institution{LinkedIn}\city{Sunnyvale}\state{California}\country{USA}}
\email{jingwu@linkedin.com}

\renewcommand{\shortauthors}{Koh et al.}

\begin{abstract}
LLM agents are becoming central to information retrieval: they issue
retrieval queries, synthesize answers from evidence, personalize
results, and increasingly serve as judges for IR evaluation. Improving
the prompts that control these agents is therefore an optimization
problem, but in applied IR settings it often looks less like blind
search and more like debugging. Engineers need to know which behavior
failed, which nearby behavior still worked, what distinguishes the two,
and whether a prompt edit improves held-out quality without introducing
regressions.

We present Contrastive Reflection, an iterative prompt-optimization
framework for agentic IR workflows. The framework starts from a
task-centric quality definition: QA agents can expose retrieval or
reasoning traces, while grading agents can expose dimension-level
scores and rationales. These structured traces are used to identify
error-anchored behavioral slices, add nearby successful examples from
the same region, and ask a Teacher LLM to propose a targeted prompt
edit. Candidate edits are accepted only when validation performance
improves, optionally subject to regression checks. We instantiate the
framework with a tree-based slice selector, but the contribution is the
contrastive reflection loop rather than the tree itself.

On a public HotpotQA retrieval-augmented QA setup, one
tree-selected contrastive repair improves held-out exact-match accuracy
from 51.4\% to 60.4\%. Failure-only and random-evidence variants also
improve, but achieve smaller held-out gains and break more previously
correct examples. A light instruction-only comparison on the same setup
places the method near modern prompt optimizers: MIPROv2 reaches 59.4\%
test accuracy and GEPA reaches 57.0\%. LinkedIn-derived grading
workflows motivate the task-centric design, but the paper's headline
evidence is the public HotpotQA study. The result is an interpretable
optimization loop for AI agents in IR, aimed at making prompt repair
more inspectable and validation-driven.
\end{abstract}



\maketitle

\section{Introduction}

LLM agents are increasingly used in information retrieval (IR)
pipelines: they generate sub-queries for multi-hop retrieval, synthesize
answers from evidence, and serve as relevance judges in place of human
annotators~\cite{thomas2023llmjudge,farzi2025criteria}. Our motivating
setting is a LinkedIn-derived grading workflow, where an LLM judge must
score whether a result is appropriate for a member or query using
multiple task-specific dimensions, rationales, and regression-sensitive
constraints. This setting shaped the method: the final grade tells a
team whether the judge matched the reference, but the dimension scores
and rationales explain \emph{how} it was wrong.

In such evaluation workflows, prompt improvement is rarely a one-shot
search problem. A prompt may perform well overall while repeatedly
making errors on a narrow class of examples: entity disambiguation in
multi-hop QA, one rating dimension in grading, or formatting behavior
for a subset of inputs. Engineers typically debug such systems by
asking five questions: which cases were wrong, which similar cases still
worked, what separates the two, what instruction would address that
contrast, and did the change help without breaking other behavior?

Contemporary prompt optimizers, including reflective and DSPy-based
methods such as GEPA and MIPROv2~\cite{gepa2025,miprov2_2025,
khattab2023dspy}, can substantially improve prompts. However, many
optimizers operate primarily through scalar validation metrics,
stochastic batches, or search over instruction-demo combinations. These
methods can find better prompts without exposing a compact explanation
of \emph{which} behavior changed. When reflection is performed on random
examples or errors alone, the prompt edit can be driven by whichever
cases happen to appear in the batch, not necessarily by a systematic
contrast between broken and working behavior.

This paper reframes prompt optimization as \emph{contrastive
reflection}. The central idea is simple: before asking an LLM to improve
a prompt, first identify a coherent slice containing errors and nearby
successes, then describe the attributes that define the slice.
Reflection becomes a targeted comparison-and-repair step rather than a
general request for better instructions. The public experiment in this
paper tests the idea on a single-predictor HotpotQA program; the
LinkedIn-derived grading setting explains the task-centric design but is
not treated as a public benchmark.

We make three contributions. First, we formalize task-centric quality
for prompt optimization: task-specific structured outputs define the
attributes used for slice discovery, regression tracking, and targeted
editing. Second, we define an iterative contrastive reflection loop:
evaluate, find contrastive slices, reflect, edit, validate, and repeat.
Third, we evaluate the idea with a public
HotpotQA~\cite{yang2018hotpotqa} study and describe how the same loop is
being refined in LinkedIn-derived grading workflows.

\section{Related Work}

\textbf{Prompt optimization.} Early prompt-tuning work optimized
continuous soft prompts~\cite{lester2021prompt}, while recent discrete
methods treat language models themselves as optimizers. OPRO and APE use
LLMs to generate and refine instructions~\cite{yang2024opro,zhou2023ape};
PromptAgent frames prompt improvement as planning~\cite{wang2024promptagent};
and APO applies gradient-inspired textual feedback~\cite{apo2023}. DSPy
provides a programming model and optimizers for multi-stage LM
programs~\cite{khattab2023dspy}; MIPROv2 searches over instructions and
demonstrations~\cite{miprov2_2025}. These methods are strong search
baselines, but they do not primarily focus on producing an interpretable
contrast between the errors and successes that motivated an edit.

\textbf{Reflection and evolutionary prompt search.} GEPA uses LLM
reflection over sampled trajectories and maintains a Pareto frontier of
candidate prompts~\cite{gepa2025}; Promptbreeder and EvoPrompt explore
related evolutionary prompt-improvement strategies~\cite{promptbreeder2023,
guo2024evoprompt}. StraGo~\cite{strago2024} is most closely related:
it shows that prompt optimization improves when the optimizer reflects
over both successes and failures rather than failures alone. Contrastive
reflection shares this intuition but differs in where the contrast comes
from. StraGo samples successes and failures from the global pool and asks
an LLM to synthesize strategic guidance across them; contrastive
reflection instead requires the successes to be drawn from the
\emph{same error-anchored behavioral slice} as the failures, so that the
contrast pins down a single, interpretable boundary that the edit must
target. A tree is one way to construct such a slice; future work could
combine contrastive evidence selection with Bayesian or evolutionary
search.

\textbf{Structured evaluation and slice discovery.} LLMs are
increasingly used as automated relevance judges in IR~\cite{thomas2023llmjudge},
and criteria-based judging decomposes relevance into explicit
dimensions~\cite{farzi2025criteria}. CheckList argues for behavioral
slices over aggregate accuracy alone~\cite{ribeiro2020checklist}. A
separate line of work focuses on \emph{discovering} such slices
automatically: SliceFinder uses decision-tree and clustering search to
find interpretable, error-heavy subgroups for model
validation~\cite{slicefinder2019}, and Domino discovers systematic
errors via cross-modal embedding clusters~\cite{domino2022}. Our slice
selector reuses the SliceFinder-style tree primitive, but the
contribution of this paper is not slice discovery itself; it is wiring
discovered slices, together with their nearby successes, into a
validation-gated reflection loop for prompt optimization. The tree is
deliberately the simplest selector that makes the contrastive evidence
interpretable to the Teacher LLM.

\section{Contrastive Reflection}

Contrastive reflection separates prompt optimization into two
questions that are often conflated. The first is \emph{where should the
optimizer look?} The second is \emph{what edit should be made?} A
general-purpose LLM can help with the second question, but it needs a
useful description of the first. Task-centric quality definitions create
that description: instead of observing only a final score, the optimizer
sees task-specific dimensions, steps, and rationales that can explain
why nearby examples diverge. The framework therefore proceeds as an
iterative loop:

\begin{enumerate}
\item Evaluate the current prompt on training and validation examples.
\item Identify error-heavy regions with interpretable shared attributes.
\item Add nearby successful examples from the same region to form a
contrastive slice.
\item Present the slice, its defining attributes, the error/success
contrast, and the current prompt section to a Teacher LLM.
\item Ask the Teacher to propose a targeted prompt edit.
\item Accept the edit only if it improves validation performance,
optionally subject to regression checks.
\end{enumerate}

In this framing, failures decide where the optimizer should look, but
successes decide what the edit must preserve. This is the main
difference between contrastive reflection and failure-only reflection:
the repair is not merely an instruction to fix bad cases, but an attempt
to infer the boundary between bad and good behavior inside the same task
region.

The slice selector in Step 2 is deliberately abstract. A decision tree
is one option, but other algorithms could be used: clustering,
embedding-nearest-neighbor search, rule mining, human-authored slices,
or learned error classifiers. The hypothesis is not that trees are
uniquely optimal. The hypothesis is that reflection improves when the
Teacher receives a coherent contrastive description instead of an
arbitrary batch of examples.

\subsection{Problem Definition}

Let $\pi_\phi$ denote an LLM program with prompt $\phi$, metric $m$,
training set $D_{tr}$, validation set $D_{val}$, and baseline prompt
$\phi_0$. Standard prompt optimization seeks a prompt that maximizes
validation performance:
\begin{equation}
\label{eq:prompt-objective}
  \max_{\phi}\; \mathbb{E}_{(x,y)\sim D_{val}}
  \left[m(\pi_\phi(x),y)\right].
\end{equation}
We denote this validation objective by $M_{val}(\phi)$.
Contrastive reflection adds an intermediate object: a slice selector
$g$ that maps current training examples and structured traces to a set
of coherent, error-anchored contrastive slices,
\begin{equation}
\label{eq:slice-selector}
  \mathcal{C}_t =
  g\!\left(D_{tr}, \{\pi_{\phi_t}(x):(x,y)\in D_{tr}\}\right).
\end{equation}
The second argument denotes the current program outputs and any
structured traces exposed by the task.
Each slice $C\in\mathcal{C}_t$ is paired with an explanation or
attribute description $a(C)$, such as a rule path, cluster summary, or
human-authored slice. A reflection operator $r$ then proposes an edit
conditioned on the current prompt and the selected contrastive evidence:
\begin{equation}
\label{eq:reflection-edit}
  \Delta\phi_t = r(\phi_t, C, a(C), E_C),
\end{equation}
where $E_C$ contains both errors and nearby successes from the same
slice. The edit is a prompt-level repair: it does not change the
training examples, but changes the instruction so the Student should
handle future examples matching the same contrastive pattern. We write
$\phi_c=\phi_t\oplus\Delta\phi_t$ for the textual operation of applying
the edit to the prompt.

The acceptance rule keeps the framework tied to held-out behavior:
\begin{equation}
\label{eq:acceptance-rule}
  \phi_{t+1} =
  \begin{cases}
  \phi_c, & \text{if } A_t(\phi_c),\\
  \phi_t, & \text{otherwise,}
  \end{cases}
\end{equation}
where $A_t(\phi_c)$ is true when
$M_{val}(\phi_c)>M_{val}(\phi_t)$ and every enabled regression check
satisfies $R_k(\phi_c,\phi_0)\le\delta_k$. Here $M_{val}$ is the
validation metric from Equation~\ref{eq:prompt-objective}. The
regression constraints $R_k$ are optional and measure protected behavior
relative to the reference prompt $\phi_0$; when disabled, acceptance
reduces to validation improvement. The main design freedom is therefore
the selector $g$: this paper studies a tree-based selector, but the
formulation permits other ways of finding useful contrastive slices.

\begin{algorithm}[H]
\caption{Contrastive Reflection}
\label{alg:cr}
\begin{algorithmic}[1]
\Require Prompt $\phi_0$, train set $D_{tr}$, validation set $D_{val}$,
metric $m$, selector $g$, reflector $r$, optional constraints $(R_k,\delta_k)$
\State $\phi_{best}\gets\phi_0$;\quad $s_{best}\gets M_{val}(\phi_0)$
\For{$t=1$ to $T$}
  \State Evaluate $\pi_{\phi_{best}}$ on $D_{tr}$ and cache traces
  \State $\mathcal{C}\gets g(D_{tr},\text{traces})$
  \State Rank slices by error density, support, and interpretability
  \ForAll{contrastive slices $C\in\mathcal{C}$}
    \State Build evidence $E_C$ from errors and nearby successes
    \State $\Delta\phi\gets r(\phi_{best},C,a(C),E_C)$
    \State $\phi_c\gets \phi_{best}\oplus\Delta\phi$
    \State $s_c\gets M_{val}(\phi_c)$
    \If{$s_c>s_{best}$ and all enabled constraints are satisfied}
      \State $\phi_{best}\gets\phi_c$;\quad $s_{best}\gets s_c$
      \State \textbf{break}
    \Else
      \State Record rejected edit for later reflection
    \EndIf
  \EndFor
\EndFor
\State \Return $\phi_{best}$
\end{algorithmic}
\end{algorithm}

\subsection{Task-Centric Contrastive Quality}

Contrastive reflection does not require every task to reduce immediately
to a single scalar signal. This is one reason the method can be more
effective than reflection over random examples: the optimizer receives
task-native evidence about \emph{how} an output is wrong, not only
whether it is wrong. For QA tasks, the program can expose intermediate
reasoning steps, retrieval queries, evidence summaries, and a final
answer; for grading tasks, it can expose dimension-level scores,
weights, and rationales. These structured outputs define the attributes
used to build slices, track regressions, and target the relevant prompt
section.

A contrastive slice is useful for reflection when it satisfies three
properties. First, it should be \emph{coherent}: errors in the slice
should differ from nearby successes for a related reason rather than
merely sharing the same final label. Second, it should be
\emph{actionable}: the shared
attributes should suggest an instruction-level intervention, such as
clarifying an entity-disambiguation rule or tightening a scoring
criterion. Third, it should be \emph{validation-relevant}: repairing the
slice should plausibly improve held-out behavior rather than memorize
isolated examples.

These criteria explain why random batches can be weak reflection inputs.
A random batch may contain a formatting error, a retrieval miss, a
reasoning error, and an ambiguous label in the same prompt. The Teacher
can still propose an edit, but the edit is forced to average over
unrelated causes. Contrastive reflection gives the Teacher a narrower
task: compare the errors and successes in this region, explain what
separates them, and write an instruction that repairs the error cases
without disturbing nearby correct cases.

In IR evaluation tasks, the acceptance step can also include
regression constraints. For example, a relevance judge may improve its
overall score while degrading a dimension that downstream ranking
experiments depend on. We therefore treat constrained acceptance as an
optional safety layer: a candidate prompt is accepted if it improves the
task metric and does not violate user-specified regression tolerances.
The core contribution, however, is the contrastive reflection loop
itself.

\section{A Tree-Based Instantiation}

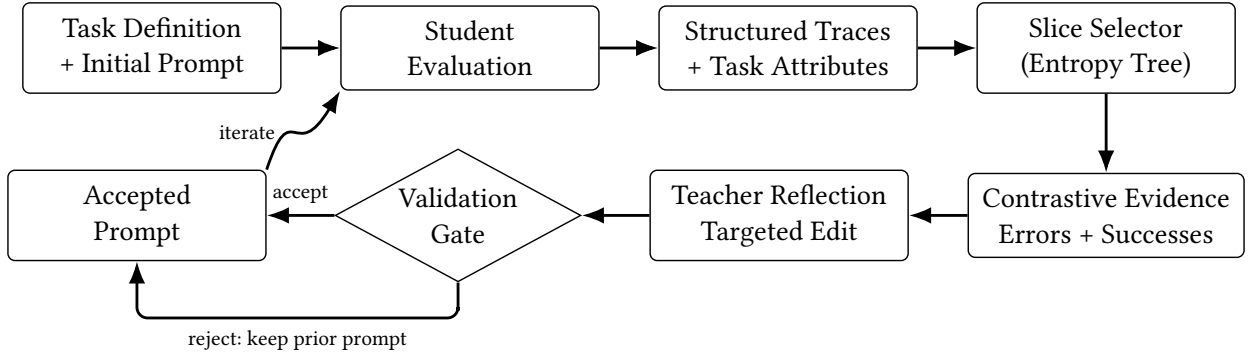
\begin{figure*}[t]
\centering
\resizebox{0.92\textwidth}{!}{%
\begin{tikzpicture}[
  box/.style={
    draw,
    rounded corners=2pt,
    align=center,
    minimum width=2.55cm,
    minimum height=0.82cm,
    inner sep=4pt,
    font=\small
  },
  gate/.style={
    draw,
    align=center,
    diamond,
    aspect=1.85,
    inner sep=1.2pt,
    font=\small
  },
  arrow/.style={-{Latex[length=2.2mm]}, thick},
  loop/.style={-{Latex[length=2.2mm]}, thick, rounded corners=3pt}
]
\node[box] (task) {Task Definition\\+ Initial Prompt};
\node[box, right=0.58cm of task] (eval) {Student\\Evaluation};
\node[box, right=0.58cm of eval] (features) {Structured Traces\\+ Task Attributes};
\node[box, right=0.58cm of features] (selector) {Slice Selector\\(Entropy Tree)};

\node[box, below=0.78cm of selector] (evidence) {Contrastive Evidence\\Errors + Successes};
\node[box, left=0.58cm of evidence] (reflect) {Teacher Reflection\\Targeted Edit};
\node[gate, left=0.67cm of reflect] (validate) {Validation\\Gate};
\node[box, left=0.67cm of validate] (prompt) {Accepted\\Prompt};

\draw[arrow] (task) -- (eval);
\draw[arrow] (eval) -- (features);
\draw[arrow] (features) -- (selector);
\draw[arrow] (selector) -- (evidence);
\draw[arrow] (evidence) -- (reflect);
\draw[arrow] (reflect) -- (validate);
\draw[arrow] (validate) -- node[above, font=\scriptsize] {accept} (prompt);
\draw[loop] (prompt.north east)
  .. controls +(0.25,0.75) and +(-0.45,-0.75) ..
  node[pos=0.34, left, font=\scriptsize] {iterate} (eval.south west);
\draw[loop] (validate.south) |- ++(0,-0.35) -| node[pos=0.25, below, font=\scriptsize] {reject: keep prior prompt} (prompt.south);
\end{tikzpicture}%
}
\caption{Contrastive Reflection optimization loop.
The Student is evaluated, task-centric traces are converted into
attributes, a slice selector identifies an error-anchored region, and a
Teacher proposes a targeted edit from contrastive evidence. The edit is
accepted only if it passes validation.}
\Description{A flow diagram showing task definition and initial prompt
flowing to student evaluation, structured traces and task attributes,
slice selection, contrastive evidence, teacher reflection, validation
gate, accepted prompt, and iteration back to student evaluation.}
\label{fig:pipeline}
\end{figure*}

We instantiate contrastive reflection with structured LLM outputs and
a shallow entropy-based decision tree. Figure~\ref{fig:pipeline} shows
the pipeline.

\subsection{Structured Outputs and Features}

The method assumes that the LLM program can expose structured outputs.
For QA tasks, these may be intermediate reasoning steps, retrieval
queries, summaries, and a final answer. For grading tasks, these may be
dimension-level scores and rationales. Structured outputs are compared
against benchmark labels or human labels when available. When those
labels are absent, a Teacher can provide diagnostic references for
debugging, but such references should be treated as optimizer evidence,
not as ground truth. These comparisons produce features such as step
correctness, score differences, reasoning similarity, format validity,
and task metadata.
Features derived directly from the final target are excluded from the
tree to avoid label leakage.

This design turns a prompt optimizer into a debugging system. Instead
of only knowing that the final answer is wrong, the algorithm can
observe that evidence selection was weak for ``who'' questions, or that
a relevance judge overscored one dimension while keeping others stable.

\subsection{Contrastive Slice Discovery with an Entropy Tree}

The proposed algorithm builds a binary decision tree over training
examples using the derived features and a correctness label. Our
implementation uses
scikit-learn's entropy criterion with CART-style binary splits. A
candidate split $\theta$ is selected by information gain:

\begin{equation}
\label{eq:information-gain}
IG(S,\theta) = H(S)
- \frac{|S_L|}{|S|}H(S_L)
- \frac{|S_R|}{|S|}H(S_R),
\end{equation}
where $S_L$ and $S_R$ are the left and right partitions induced by
$\theta$, and $H(S)=-\sum_{c\in\{0,1\}} p_c(S)\log_2p_c(S)$ is the
entropy of the correctness labels in node $S$, with $0\log 0=0$.

The tree is useful because each root-to-leaf path is a compact rule
that can be inserted directly into a reflection prompt. In this
instantiation, leaves are selected because they are error-heavy, but
their evidence can include both error cases and successful cases from
the same path. A selected leaf is therefore not just a bag of errors; it
is a contrastive slice, such as long predictions in examples where the
answer appears in the retrieved context. This makes the subsequent
reflection step more concrete.

\subsection{Reflection and Acceptance}

For each selected leaf, the algorithm gives the Teacher LLM four inputs:
the tree path, a small set of error examples, a small set of successful
examples from the same region when available, and the relevant prompt
section. The Teacher then repairs the observed error pattern by
inferring an instruction-level distinction: what rule would make the
Student treat the error cases more like the successful cases without
breaking the successes? The proposed edit is applied to the prompt, not
to the examples. The Student program is then evaluated on validation
examples; rejected edits are recorded so later reflection calls avoid
repeating the same ineffective strategy.

\section{Experiments}

We evaluate on HotpotQA with a lightweight retrieval-augmented QA
program. For each question, the program retrieves the top four context
paragraphs by lexical overlap from the provided HotpotQA context and
asks a GPT-4.1-mini Student model for a short answer. This is a
single-predictor public study, not an evaluation of every
agent workflow that motivated the framework. We use 2{,}000 training
examples for slice discovery, 500 validation examples for acceptance,
and 500 held-out test examples for reporting. All
contrastive-reflection variants use the same split, retrieval code,
Student model, Teacher model, and exact-match metric.

\begin{table}[t]
\centering
\caption{HotpotQA exact-match results for one prompt-repair step.}
\label{tab:hotpot}
\footnotesize
\setlength{\tabcolsep}{2.5pt}
\begin{tabular}{lccccc}
\toprule
Method & Val & Test & $\Delta$ Test & 95\% CI & $p$ \\
\midrule
Baseline prompt & 56.0 & 51.4 & --- & --- & --- \\
\textbf{Tree contrastive} & \textbf{64.0} & \textbf{60.4} & \textbf{+9.0} & [6.0, 12.0] & $<.0001$ \\
Tree failure-only & 57.4 & 54.6 & +3.2 & [0.4, 6.0] & .0402 \\
Random contrastive & 63.4 & 59.0 & +7.6 & [4.4, 10.8] & $<.0001$ \\
\bottomrule
\end{tabular}
\end{table}

Table~\ref{tab:hotpot} shows that a single tree-selected contrastive
repair improves held-out exact match from 51.4\% to 60.4\%. The
failure-only and random-evidence variants also improve, which indicates
that answer-form precision is a strong repair opportunity in this setup
and that any reflection-based method is likely to find a generic
answer-form fix. The \emph{distinct} signal for tree-contrastive in
this setup is therefore not the +1.4 pp test margin over random
contrastive (well within the bootstrap interval), but the cleaner
repair profile: tree-contrastive fixed 54 held-out examples while
breaking 9, versus 52 fixed / 14 broken for random contrastive and 35
fixed / 19 broken for failure-only. This is the behavior the method is
designed to encourage: use failures to choose where to look, use nearby
successes to define what must be preserved, and accept only after
validation improves. We report exact McNemar $p$-values for paired
exact match and bootstrap 95\% confidence intervals over test-set
examples. We use \emph{fixed} to mean a held-out example that was
incorrect under the baseline prompt and correct after the edit;
\emph{broken} means the reverse.

The selected tree leaf also made the repair mechanism inspectable. Its
failures were mostly answer-form near misses, such as returning a
sentence or descriptor instead of the shortest exact answer. Nearby
successes showed the desired behavior: concise entity or year answers
with no extra phrasing. The accepted Teacher edit can be summarized as
one instruction: \emph{return only the exact final short answer as found
in the context, with no explanation or added information}. The edit
fixed 54 held-out examples while breaking 9.

\begin{table}[t]
\centering
\caption{Ablation study for the same one-step HotpotQA setup.}
\label{tab:ablation}
\footnotesize
\setlength{\tabcolsep}{4pt}
\begin{tabular}{lcccc}
\toprule
Variant & $\Delta$ Val & $\Delta$ Test & Fixed & Broken \\
\midrule
\textbf{Tree contrastive} & \textbf{+8.0} & \textbf{+9.0} & \textbf{54} & \textbf{9} \\
Tree failure-only & +1.4 & +3.2 & 35 & 19 \\
Random contrastive & +7.4 & +7.6 & 52 & 14 \\
\bottomrule
\end{tabular}
\end{table}

Table~\ref{tab:ablation} ablates two questions: whether the evidence
comes from an interpretable tree slice, and whether the Teacher sees
successes as well as errors. The contrastive tree slice gives the best
validation and held-out gains while breaking the fewest examples.
Random contrastive evidence also improves, but its lower test gain and
larger broken count suggest that contrastive evidence is most useful
when the selected cases share a coherent behavioral pattern.

The validation gate also matters. The first repair is accepted because
validation exact match rises from 56.0\% to 64.0\%. A second
tree-selected repair targets a different failure mode, imprecise short
answers that omit qualifiers, but the edit over-corrects: validation
drops to 58.2\%, fixing only 7 validation examples while breaking 36.
The algorithm therefore keeps the first accepted prompt, and the
held-out test result remains 60.4\%.

\begin{table}[t]
\centering
\caption{Light optimizer sanity check on the same single-predictor
HotpotQA split. DSPy baselines use instruction-only settings; GEPA uses
a 100-example optimizer validation slice.}
\label{tab:optimizer}
\footnotesize
\setlength{\tabcolsep}{4pt}
\begin{tabular}{lccc}
\toprule
Method & Val & Test & $\Delta$ Test \\
\midrule
Baseline prompt & 56.0 & 51.4 & --- \\
GEPA-light & 61.2 & 57.0 & +5.6 \\
MIPROv2-light & 62.2 & 59.4 & +8.0 \\
\textbf{Tree contrastive} & \textbf{64.0} & \textbf{60.4} & \textbf{+9.0} \\
\bottomrule
\end{tabular}
\end{table}

Table~\ref{tab:optimizer} is a sanity check against modern optimizers,
not a definitive optimizer ranking. We intentionally use the same
2{,}000/500/500 single-predictor setup from Table~\ref{tab:hotpot} and
light instruction-only DSPy settings so the comparison is tied to the
headline public experiment. The configurations are not cost-matched:
GEPA's reflection budget is a 100-example optimizer validation slice
rather than the 500-example slice contrastive reflection accepts on,
and MIPROv2 is run in instruction-only mode rather than its full
instruction-and-demonstration search. Under these caveats, MIPROv2 is
close on held-out accuracy (the 1.0 pp gap sits within the bootstrap
interval reported in Table~\ref{tab:hotpot}), so we read the result as
\emph{competitive, not better}: contrastive reflection lands in the
same band as a modern optimizer while also producing a compact account
of the failure slice, the nearby successes, and the accepted edit.
GEPA improves over the baseline as well, but its selected prompt
generalizes less strongly to the held-out split in this single run.
Cost-equivalent benchmarks for multi-stage agent programs remain
future work.

\subsection{LinkedIn-Derived Grading Workflows}

The LinkedIn-derived grading setting remains important to the narrative
because it is where the task-centric quality definition arose. In these
workflows, an LLM judge assigns a final 0--4 grade from structured
evidence while also producing dimension-level scores and rationales. The
final grade provides the scalar metric, but the dimensions and rationales
explain \emph{how} the judge is wrong and which instruction should be
edited.

Refinement of the method on this setting is ongoing, so we do not report
a formal table in this workshop version. Internal development has shown
approximately a +12 percentage-point gain on held-out grading accuracy,
with useful prompt repairs typically reached in 3--5 optimization
epochs. We treat this as motivating evidence rather than a public
benchmark claim because the examples, rubrics, and labels are not
distributable. The lesson we carry into the public HotpotQA study is
about speed and inspectability: structured slices let engineers identify
which grading dimension is failing, compare it with nearby correct
judgments, and make a targeted grading-rule edit without waiting for
many rounds of unconstrained prompt search.

\section{Discussion}

The central lesson is that prompt optimization can be treated as a
debugging loop. Scalar validation metrics remain necessary, but they do
not tell the optimizer where to reason. Contrastive reflection adds a
middle layer: find a coherent slice where errors and successes diverge,
describe the contrast, and ask for a repair that targets that slice.
The HotpotQA result should therefore be read as evidence for the
mechanism, not as proof that a decision tree is always the best
optimizer.

The decision tree is a practical instantiation, not the final word.
Trees are attractive because they produce human-readable paths and
natural reflection prompts. Other slice selectors may work as well,
and future work should compare trees against clustering, embedding-based
retrieval, rule mining, and human-authored slices. More seeds and more
tasks are also needed before making strong claims about optimizer
rankings.

The approach has limitations that future work should address. It
requires structured outputs, which add token overhead and may require
prompt/interface design. Teacher-generated diagnostic references can
encode Teacher errors or biases and should not be treated as human
labels. The public HotpotQA study is intentionally narrow: it tests
one-step prompt repair on one dataset, a single-predictor QA program,
and a single random seed; variance is reported only via bootstrap
intervals on test predictions, not across reruns of the full
optimization loop. Two specific gaps follow from this scope. First,
because only one edit is ultimately accepted on the public benchmark,
the experiments mostly exercise the contrastive \emph{step} rather than
the iterative loop that Algorithm~\ref{alg:cr} centers on. Second, the
regression-constraint machinery in
Equation~\ref{eq:acceptance-rule} is defined and used in the
LinkedIn-derived grading setting but is not exercised on the public
HotpotQA setup, where validation-only acceptance is sufficient. Broader
task coverage, multi-stage agent programs, multi-seed reruns of the
full loop, and a public regression-constrained study remain future
work. The LinkedIn-derived grading observations are included only as
motivating context because the data and rubrics are not distributable;
they should not be read as a fully reproducible public benchmark.

\paragraph{Reproducibility.}
All headline HotpotQA runs use the same 2{,}000/500/500 split,
exact-match scoring, retrieval code, and GPT-4.1-mini Student/Teacher
models. The implementation records prompts, traces, slice reports, token
usage, and before/after comparisons for each table row.

\section{Conclusion}

Contrastive Reflection frames prompt optimization as an iterative loop
of task-centric quality definition, slice discovery, reflection,
targeted editing, and validation. Failures identify where the prompt is
weak; nearby successes reveal the behavior that the repair should
preserve. Our proposed algorithm instantiates this loop with structured
outputs and tree-based contrastive slice discovery. On HotpotQA, it
improves held-out exact-match accuracy from 51.4\% to 60.4\% in one
repair step. Ongoing LinkedIn-derived grading work motivates the same
design goal: reach useful prompt repairs quickly while preserving an
inspectable account of which task dimension changed. The broader agenda
is to make prompt optimization less like blind search and more like
inspectable, repeatable debugging.

\bibliographystyle{ACM-Reference-Format}
\bibliography{refs}

\end{document}